# Frog-Snake prey-predation Relationship Optimization (FSRO) : A novel nature-inspired metaheuristic algorithm for feature selection

Hayata SAITOU[1] · Harumi HARAGUCHI[1]

## Abstract

Swarm intelligence algorithms have traditionally been designed for continuous optimization problems, and these algorithms have been modified and extended for application to discrete optimization problems. Notably, their application in feature selection for machine learning has demonstrated improvements in model accuracy, reduction of unnecessary data, and decreased computational time. This study proposes the Frog-Snake prey-predation Relationship Optimization (FSRO) algorithm, inspired by the prey-predation relationship between frogs and snakes for application to discrete optimization problems. The algorithm models three stages of a snake's foraging behavior "search", "approach", and "capture" as well as the frog's characteristic behavior of staying still to attract and then escaping. Furthermore, the introduction of the concept of evolutionary game theory enables dynamic control of the search process. The proposed algorithm conducts computational experiments on feature selection using 26 types of machine learning datasets to analyze its performance and identify improvements. In computer experiments, the proposed algorithm showed better performance than the comparison algorithms in terms of the best and standard deviation of fitness value and Accuracy. It was also proved that dynamic search control by evolutionary game theory is an effective method, and the proposed algorithm has the ability of a well-balanced search, achieving the two objectives of improving accuracy and reducing data.

✉ Hayata SAITOU
22nm717g@vc.ibaraki.ac.jp

Harumi HARAGUCHI
harumi.haraguchi.ie@vc.ibaraki.ac.jp

[1] Graduate school of Science and Engineering, Ibaraki University, 4-12-1 Nakanarusawa, Hitachi, 3168511 Ibaraki, Japan

## 1. Introduction

Meta-heuristics is optimization algorithms that are not dependent on a specific problem. Since they require less computational capacity and can efficiently solve a wide variety of complex optimization problems, they have better performance than deterministic algorithms [1]. Therefore, it has become prevalent in many application fields over the past few decades.

Meta-heuristics can be categorized into two groups: Single-based and Population-based. Single-based includes Simulated Annealing (SA) [2] and Tabu search (TS) [3]. Population-based can be further subdivided into various categories. Evolutionary-based in this category are Genetic algorithm (GA) [4] and Differential Evolution (DE) [5], Swarm-based, called Swarm intelligence algorithm, include Particle Swarm Optimization (PSO) [6], Grey Wolf Optimization (GWO) [7], Harris Hawk Optimizer (HHO) [8]. Human-based include Harmony Search (HS) [9], Teaching-Learning-Based Optimization (TLBO) [10], Bio-based include Earthworm Optimization Algorithm (EOA) [11], Virus colony search (VCS) [12]. Physics-based include Atom Search Optimization (ASO) [13], Equilibrium optimizer (EO) [14], Math-based include Sine Cosine Algorithm (SCA) [15], Arithmetic Optimization Algorithm (AOA) [16]. In particular, swarm intelligence algorithms have developed since the proposal of PSO in 1995, with the development of improved methods and theoretical research on their performance. Many swarm intelligence algorithms have been proposed and designed inspired by group migration, foraging, survival, and defense behaviors.

Especially, Grey Wolf Optimization (GWO) is inspired by the hunting behavior of wolves. GWO has the advantages of high exploitation and fast convergence, however, it has the disadvantages of low exploration and stagnation to local optima [17]. In the past, the author modified and applied the Oppositional Based Learning (OBL) method which is effective in improving the exploration and the ability to escape from local optima to improve GWO. As a result, the search performance was improved on some benchmark problems. However, high exploitation and fast convergence were lost, so the parameter of OBL method was adjusted to maintain the previous search performance. This algorithm was proposed as Improved Mutation Oppositional Based Learning Grey Wolf Optimization (IMOGWO) and showed the best results in computer experiments using 28 benchmark functions [18][19]. In most situations, it is very difficult to set an appropriate balance between exploration and exploitation in swarm intelligence algorithms. Therefore, it is necessary to design algorithms that can control the search phase.

Most swarm intelligence algorithms are designed for continuous optimization problems. Therefore, binary versions of swarm intelligence algorithms are proposed, for instance, binary PSO (BPSO) and binary GWO (BGWO) [20][21]. These algorithms can be applied to the feature selection in machine learning, defined as a binary optimization problem. The Wrapper method, a type of feature selection, can improve the accuracy of the classification model and remove unnecessary data however, it is computationally expensive. For a feature vector size $N$, the feasible solution space is $2^N - 1$. Therefore, the application of swarm intelligence algorithms is effective in finding a better solution in a huge space.

A binary version of swarm intelligence algorithms shows excellent performance in solving binary optimization problems. However, the performance of algorithms converted from continuous to discrete versions differs depending on various factors, such as the setting of the appropriate parameters and transfer functions, the classifier used, and the combination of algorithms adopted for hybridization [22]. It is difficult to find experimentally the best settings for these various factors because of experimental time and costs. Therefore, the development of swarm intelligence algorithms for application to discrete problems and providing researchers with a basic framework that can be easily improved and applied enables contributing to this field.

In this study, a novel swarm intelligence algorithm for application to discrete optimization problems, which is called Frog-Snake prey-predation Relationship Optimization (FSRO), is proposed inspired by the prey-predator relationship between frogs and snakes in nature. This algorithm models the search, approach, and capture phases of the snake's foraging behavior, as well as the characteristic behavior of the frogs in confrontation, involving "to stand still, attract, and then flee" [23].

The proposed algorithm introduces the concept of evolutionary game theory for dynamic control of the search. Evolutionary game theory (EGT) is a field that uses game theory to analyze biological behavior and strategy, and evolutionary phenomena. It has been applied not only to biology but also to computer science, including optimization techniques, and has contributed to solving various problems. The proposed algorithm dynamically controls the search by designing mutation operations with evolutionary stable strategy and adjustment of share of population with replicator dynamics, which are important concepts in EGT.

26 datasets from UCI Machine Learning Repository [24] and ASU Feature Selection Repository [25] are used in the experiments to verify the effectiveness of the proposed algorithm. The datasets are selected to have small to large number of features.

The rest of this paper is organized as follows. Section 2 presents related work. Section3 describes the evolutionary game theory (EGT). Section 4 describes the behavior of frogs and snakes in nature and, the model of the proposed algorithm named FSRO. In section 5, the experimental results are presented and discussed. Finally, conclusions are given in section 6.

## 2. Related work

Binary version of PSO was proposed in 1997 based on the original PSO by Kennedy and Eberhart [19]. In this study, the continuous search space is mapped to the binary search space through a sigmoidal function as a transfer function, allowing PSO to search in the binary space. In this study, the position update equation is modified, and velocity is replaced as a value that affects the binary selection probability of the solution. Recently, many swarm intelligence algorithms converted from continuous to binary versions have been proposed. A list of major binary versions of the swarm intelligence algorithm is shown in Table 1.

Table 1 : The list of main binary versions

| Algorithm | year |
|---|---|
| Binary Particle Swarm Optimization [20] | 1997 |
| Binary Ant Algorithm [26] | 2007 |
| Binary Bat Algorithm [27] | 2013 |
| Binary Grey Wolf Optimization [21] | 2015 |
| Binary Dragonfly algorithm [28] | 2017 |
| Binary Harris Hawks Optimization [29] | 2019 |
| Binary Whale Optimization Algorithm [30] | 2020 |

Guo et al. identified the most effective transfer function that improves search performance in BPSO among S-shaped, such as the sigmoid function, V-shaped, and Z-shaped transfer functions [31]. Experimental results for 12 different transfer functions on standard benchmark functions showed that the best search performance was achieved when using a Z-shaped transfer function.

Kulandaivel et al. demonstrated the effectiveness of S-shaped transfer functions in binary African vulture optimization algorithm (BAVOA) [32]. Dinar et al.

demonstrated the effectiveness of V-shaped transfer functions in Salp Swarm Algorithm (SSA) [33].

As described above, the binary versions of algorithms are applied to various fields such as breast cancer (BC) prediction, bearing fault diagnosis, medical diagnosis, dynamic complex protein detection, unit commitment problem (UC), text categorization [34-39].

Rajamohana et al. proposed the hybrid approach of improved BPSO (iBPSO) and shuffled frog leaping algorithm (SFLA), named iBPSO_SFLA [40]. In this study, the experimental results compared with other feature selection algorithms showed that iBPSO_SFLA offered high classification accuracy for feature selection and is efficient. Especially, it successfully classified the reviews available on the internet into fake reviews and trusted reviews.

RANYA et al. proposed the hybrid approach of Binary Grey Wolf Optimization (BGWO) and Harris Hawks Optimizer (HHO), named HBGWOHHO [41]. This performance was improved by replacing the BGWO exploration phase with HHO exploration phase. Experimental results with 18 standard data sets showed that HBGWOHHO has better performance than the BGWO. It achieved high accuracy and a small size of selected feature in a short computational time compared to the other algorithms such as Binary GA (BGA) and BPSO.

Nabil Neggaz et al. proposed the hybrid approach of SSA and Sine Cosine Algorithm (SCA), named ISSAFD [42]. In this algorithm, the exploration phase of SSA was improved, and new operators were applied to ensure the diversity of the population to avoid stagnation in a local optimum. Experimental results on 20 datasets, including a high-dimension dataset of about 10000 dimensions, showed that ISSAFD performed better than other algorithms in terms of accuracy, true positive patterns, true negative patterns, and size of selected features.

Recently, a new optimization approach called hyper-heuristics has attracted researchers' attention. This is a field that automatically selects or generates algorithms to be applied during execution based on the characteristics of the solution space and the search stage. By combining the strengths of multiple algorithms and compensating for their weaknesses, it enables effective search.

A.Charan et al. proposed the hyper-heuristic approach based on various versions of GA, named Multi-objective Hyper-heuristic Evolutionary Algorithm (MHypEA) [43]. This is an approach that adaptively selects effective algorithms for solving problems from 12 different GAs with different parent selection, crossover operators, and mutation. Based on the weights assigned to each GA, the search algorithm was selected using roulette-wheel selection.

Rehab et al. proposed the hyper-heuristic method named CODMSP, which stands for Chaotic map and Opposite-based learning DE and MH technique for Selection Paradigm [44]. This approach optimized various components by using DE to identify the appropriate combination of the use of chaos maps in generating the initial population, the ratio of the population applied to OBL, and the Meta-Heuristics to search the solution space. Experimental results on standard machine learning datasets show that it performs better than state-of-the-art swarm intelligence algorithms for feature selection in terms of accuracy.

From these related works, it is found that many researchers have faced challenges in determining appropriate parameter settings, selecting suitable transfer functions for binarization, and effectively utilizing different heuristics. However, discovering the optimal configurations among these factors is not an easy task and is often needed with experimental costs and time-consuming processes.

Cédric Leboucher et al. proposed an improved PSO using evolutionary game theory (EGT), named Combined-Evolutionary Game PSO (C-EGPSO) [45]. The parameters in the position and velocity update equations for each particle are adjusted by EGT based on their previous solutions to optimize the search direction. This improvement was successful in maintaining population diversity and adapting the search phase to the shape of the benchmark function.

Ziang Liu et al. proposed an improved PSO using EGT based on an ensemble approach, named Strategy Dynamics PSO (SDPSO) [46]. This consists of four improved versions of PSO, dynamically selecting effective algorithms for obtaining better solutions among them by replicator dynamics. Experimental results on the CEC2014 benchmark problem showed the superior performance of SDPSO compared to the other algorithms. However, it was indicated that the effects of incorporating other improved versions of PSO need to be verified and discover the optimal configuration.

## 3. Evolutionary Game Theory

In this section, the evolutionary game theory (EGT) is described, which is applied to control of dynamic search of the proposed algorithm. Evolutionary game theory is an extension of game theory to biology. Since Maynard Smith proposed it in 1970s, it is applied to economics, finance, and various area [47]. As shown in related work, it is also applied to optimization techniques. Evolutionary stable strategy (ESS) and replicator dynamics are two important concepts in evolutionary game theory. The details of them are given in the following section.

### 3.1. Evolutionary Stable Strategy (ESS)

The concept of equilibrium in evolutionary game theory is defined as an evolutionary stable strategy. In a population consisting most members of the population playing strategy A, if individuals playing other strategies other than A cannot invade into the population, strategy A is called an evolutionary stable strategy. Defining the fitness of an individual playing strategy X, when ?? an individual playing strategy Y as $E(X,Y)$, Eq. (3.1-3.2) is established.

$$E(X,A) < E(A,A) \quad (3.1)$$
$$E(X,X) < E(A,X), in\ case\ E(X,A) = E(A,A) \quad (3.2)$$

### 3.2. Replicator Dynamics

Replicator dynamics is a concept for determining changes in share of strategy in a population. The idea that strategies with high payoff increase the individual share is represented by the following Eq. (3.3)

$$x_h(t+1) = x_h(t) + x_h(t) \times [u_h(e^h, x) - u(x, x)] \quad (3.3)$$

where $x_h(t)$ is share of strategy $h$ in generation $t$ and, $u_h(e^h, x)$ is the payoff of the individual playing strategy $h$. $u(x, x)$ is the average of fitness in population, represented in Eq. (3.4).

$$u(x,x) = \sum_{h=1}^{k} x_h u(e^h, x) \quad (3.4)$$

## 4. The proposed algorithm

In this section, a concept of the predator-prey relationship between frogs and snakes in nature is described. Then, the behavior of frogs and snakes is modeled as an optimization algorithm to apply to binary problems. An introduction of evolutionary game theory is also described.

### 4.1. Frog and snake relationship in nature

Predators and prey have evolved to refine their strategy to win each other in the process of evolution. A predator-prey relationship between frogs and snakes is known as one type of this relationship. A snake have three phases of predatory behavior "search", "approach" and "capture". In each phase, frogs exhibit the characteristic behavior of "standing still, attracting, and then running away" [23]. The details of each phase are described as follows.

#### 4.1.1. Search behavior

Search behavior is the phase that the snake searches for the habitat of the frog. At long distances, the immobility of frogs has the effect of allowing it to avoid being noticed by the snake and becoming a target for predation. At short distances, the frog's sudden movement from a stationary position has the effect of surprising the predator and allowing it to avoid predation. In the study by Nishiumi et al., the percentage of snakes that preyed at long and short distances was obtained for each frog behavior (Table 2) [48].

Table 2 : Percentage of snakes that preyed

| Distance of frogs and snakes | Behavior of frogs | |
|---|---|---|
| | Motionless | Moving |
| Long Distance (40cm~80cm) | 0% | 100% |
| Short Distance(0cm~10cm) | 89% | 70% |

#### 4.1.2. Approach behavior

Approach behavior is the phase that the snake notice the frog and moves closer to narrow the distance between it. The immobility of frogs has the effect of leading snakes to change their targets [49]. The immobility of frogs prevents other nearby frogs from detecting the presence of snakes. As a result, the snake changes the target to the nearby frog, allowing the initial target to escape. Therefore, it is an effective behavior not only in the tactics between predators and prey but also between prey themselves.

#### 4.1.3. Capture behavior

Capture behavior is the phase that the snake reaches near the frogs and attempt to bite and capture it. The immobility of frogs has the effect of evading predation by strategically choosing when to act. If the snake moves first and the frog can predict its movement and then move second, it is possible to escape. At a close distance, however, the frog cannot avoid the snake's initial capture, so it is effective to move first. Therefore, by stationary, the frog estimates the distance from the snake and chooses the most effective action to avoid predation. In the study by Nishiumi et al., distance between frog and snake at first or second movement was obtained (Fig 1). In this table, CD represents critical distance for evading an initiated strike, SID represents strike initiation distance, FID represents flight initiation distance. In addition, probability of successful escape by order of behavior were obtained (Fig 2) [50].

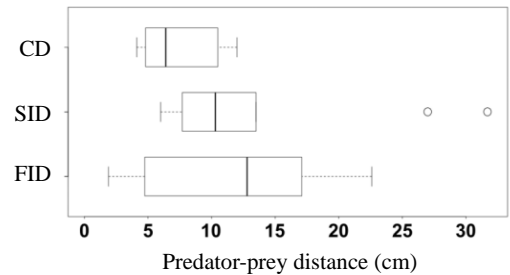

Fig 1 : Distance at first or second movement

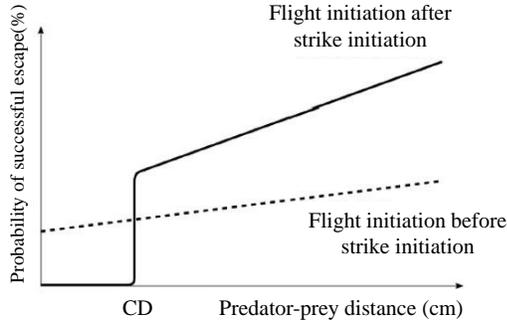

Fig 2 : Probability of successful escape

### 4.2. Model of frog and snake relationship

In this subsection, the mathematical models of the predator-prey relationship between frogs and snakes in nature are described.

#### 4.2.1. Initialization

The number of populations of each frog group and snake group is set to $N/2$, where $N$ is population size, generating a $D$-dimensional solution with each index randomly set to 0 or 1. The initialization equation is shown Eq. (4.1-4.2)

$$x_F = (x_1, x_2, x_3, \ldots, x_{N/2}), x_i \in \{0,1\}, i \in \{1,2,3,\ldots,N/2\} \quad (4.1)$$
$$x_S = (x_1, x_2, x_3, \ldots, x_{N/2}), x_i \in \{0,1\}, i \in \{1,2,3,\ldots,N/2\} \quad (4.2)$$

where $x_F$ is the solution of frogs and $x_S$ is the solution of snakes.

#### 4.2.2. Basic solution update

In the snake group, the solution is updated by two-point crossover, which has the performance of exploration. The crossover point is randomly selected from the $D$-dimensional solution solutions. Fig 3 shows the operation of a two-point crossover.

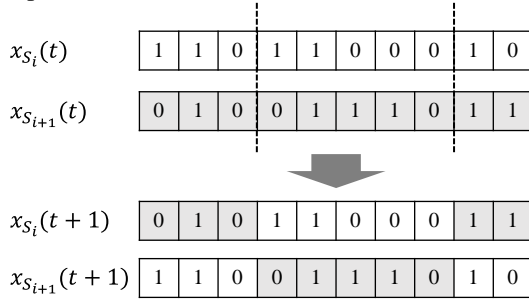

Fig 3 : Two-point crossover

In the frog group, the solution is updated by uniform crossover, which has the performance of exploitation. The index of the crossover point are chosen at 50%. Fig 4 shows the operation of an uniform crossover

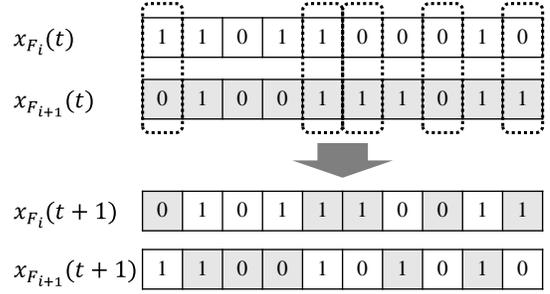

Fig 4 : Uniform crossover

#### 4.2.3. Search phase

In the search phase, the behavior of the frogs is determined based on the results of uniform crossover. The number of indexes changed by uniform crossover and the number of indexes unchanged are obtained. The behavior of the frogs is determined according to the following rules.
- if [Changed index]>[Unchanged index], Moving
- if [Unchanged index] ≤[Changed index], Motionless

Figure 5 shows the process of determining the behavior of frogs.

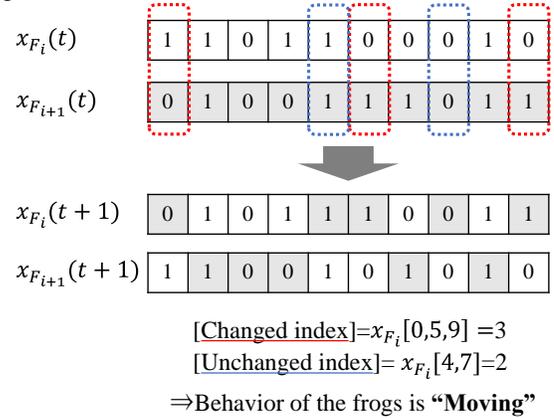

[Changed index]=$x_{F_i}[0,5,9]$ =3
[Unchanged index]= $x_{F_i}[4,7]$=2
⇒Behavior of the frogs is **"Moving"**

Fig 5 : Process of determining the behavior of frogs

#### 4.2.4. Approach phase

In the approach phase, the changed indexes upon successful predation are determined in the frog group. The indexes replaced by the uniform crossover are randomly selected. Predation points are determined by the following two rules.
- Escape behavior
  If a randomly selected index is a changed index, the selected index is set to the predation point.
- Immobile behavior
  If a randomly selected index is an unchanged index, the predation points are set to the all indexes after (before) the crossover point of the nearest uniform crossover to the selected index.

Figure 6 shows the process of determining the predation point.

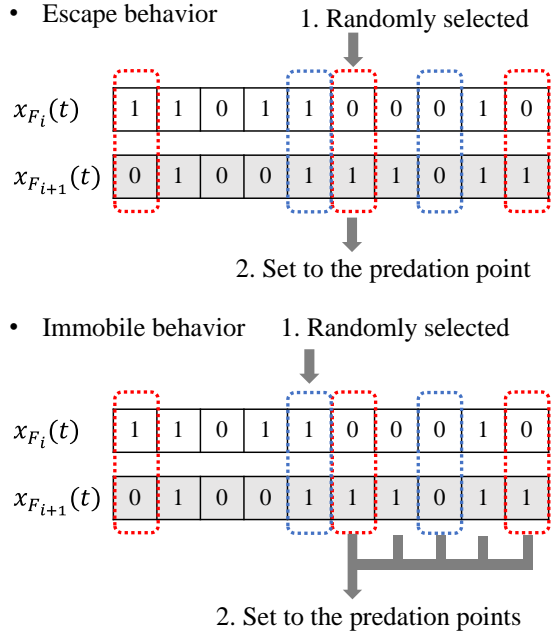

Fig 6 : Process of determining the predation point

### 4.2.5. Capture phase

In the capture phase, the distance between the frogs and snakes and the first or second move of both are determined. Then, the elements set as predation points are inverted between 0 and 1 when predation is successful. The distance between frogs and snakes is determined by the number of matches between randomly selected frog and snake solution elements. The distance equation is shown Eq. (4.3)

$$FrogSnakeDis = MaxDis \times \frac{dim - NumOfMatct}{dim} \quad (4.3)$$

where $MaxDis$ is constant of the max distance between frogs and snakes. $dim$ is the dimension of the solution (total features) and $NumOfMatct$ is the number of matched elements.

The first and second moves are determined as in equation (4.4) below.

$$FrogOrder = \begin{cases} First, if\ FrogSnakeDis \leq DicisionDis \\ Second, if\ FrogSnakeDis > DicisionDis \end{cases} \quad (4.4)$$

where $DicisionDis$ is constant for determining the first and second moves.

The predation avoidance rate is represented by the following Eq. (4.5-4.6)

$$FirstAvoidanceRate = \frac{w_1 \times FrogSnakeDis + d_1}{100} \quad (4.5)$$

$$SecondAvoidanceRate = \frac{w_2 \times FrogSnakeDis + d_2}{100} \quad (4.6)$$

where $w_1$ and $w_2$ are constants for determining the predation avoidance rate, $d_1$ and $d_2$ are the minimum predation avoidance rate.

### 4.3. Dynamic control by EGT

In this subsection, the introduction of evolutionary game theory into the proposed algorithm is described.

### 4.3.1. Population adjustment

Replicator dynamics dynamically adjust the number of population of frogs and snakes. The payoff in evolutionary game theory is calculated based on the improvement of the fitness value, and this value is used to adjust the number of individuals. The improvement of the fitness value and its average is given by the following Eq. (4.7-4.8).

$$\Delta F_h(t) = \Delta F_h(t) + \left(F(X_i(t)) - F(X_i(t-1))\right) \quad (4.7)$$

$$\Delta \bar{F}_h(t) = \frac{\Delta F_h(t)}{NumOfAgent_h} \quad (4.8)$$

where $h \in \{1,2\}$, $NumOfAgent_h$ is the number of each population.

The equation that converts the improvement of fitness value to payoff is given by Eq. (4.9)

$$u_h(t) = \frac{\Delta \bar{F}_h(t)}{\sum_{h=1}^{k} \Delta \bar{F}_h(t)} \quad (4.9)$$

where $u_h(t)$ is payoff of the number of each population and $u_h(t) \in [0,1]$.

The equations for calculating the percentages of each population in the next iterations are shown in Eq. (4.10-4.11).

$$x_h(t+1) = x_h(t) + x_h(t) \times \left(u_h(t) - \bar{u}(t)\right) \quad (4.10)$$

$$\bar{u}(t) = \sum_{h=1}^{k} x(t) \times u_h(t) \quad (4.11)$$

where $x_h(t+1)$ is the percentages of each population in the next iterations and $\bar{u}(t)$ is average of the all payoff.

By these calculations, the number of each population is dynamically controlled. Thus, the number of population of the group that improved the fitness value increases in the next generation, while the number of population of the group that did not improve the fitness value decreases.

### 4.3.2. Mutation Operation

Evolutionary stable strategy (ESS) means that the population of one group occupies the majority of the population. Mutation operation is performed to maintain diversity of the population. If the number of population in a group becomes 2 or fewer, the best solution until current iteration is added to that group.

### 4.4. Pseudocode of FSRO

The pseudocode of FSRO is presented as follows.

**Begin**
  Initialize
  **While** (t<Max Iteration)
    **Update** the solution of frogs and snakes by crossover
    **Calculate** the changed and unchanged index (Search)
    **Determine** the predation points (Approach)
    **Invert** predation points at 0 and 1 (Capture)
    **for** each search agents
      Calculate the fitness of all search agents
    **End for**
    **Calculate** improvement of fitness value
    **Adjustment** population by replicator dynamics
    **Mutation** by evolutionary stable strategy
    t = t + 1
  **End while**
  **Return** best fitness
**End**

## 5. Experimental result and discussions

In this section, computer experiments on feature selection using machine learning datasets are conducted to evaluate the performance of the proposed algorithm. Various experimental conditions and results are described and discussed.

### 5.1. Experimental condition

Details of the various experimental conditions are described in this section.

#### 5.1.1. Fitness equation

The fitness equation used in the proposed algorithm to evaluate each agent is as shown in Eq. (5.1)

$$Fitness = \alpha E_R(D) + \beta \frac{|R|}{|C|} \quad (5.1)$$

where $E_R(D)$ is error rate of the classification, $R$ is length of selected feature, $C$ is the total number of features. $\alpha$ and $\beta$ are constants corresponding to the importance of classification accuracy and feature reduction. $\alpha \in [0,1]$ and $\beta = 1 - \alpha$, $\alpha = 0.9$ in this experiment.

#### 5.1.2. Classification method

In this experiment, KNN (K-nearest neighbor) is used as a classifier to check the goodness of the selected feature. This is a supervised learning algorithm that performs classification of unknown instances based majority voting of k-nearest neighbor categories. In this experiment, the dataset is divided into 80% training data and 20% test data.

#### 5.1.3. Evaluation criteria

Seven evaluation criteria used to evaluate the effectiveness of proposed algorithm are the mean, best, worst, standard deviation of fitness, average accuracy, average reduction rate of features, computational time. The experiment is run 30 times to ensure stability and statistical significance of the results. The equations of each criterion are shown in Eq. (5.2-5.8).

- Mean of fitness

$$meanFit = \frac{1}{M}\sum_{k=1}^{M} Fit(x_k) \quad (5.2)$$

where $M$ is the number of runs. $k$ is the population number, $k \in \{1,2,3,\dots,N\}$.

- Best of fitness

$$bestFit = \min(fit(x_k)) \quad (5.3)$$

- Worst of fitness

$$Worst\ Fitness = \max(fit(x_k)) \quad (5.4)$$

- Standard deviation of fitness

$$stdFit = \sqrt{\frac{1}{M}\sum_{k=1}^{M}(fit(x_k) - Average\ Fitness)^2} \quad (5.5)$$

- Average accuracy

$$AveAcc = \frac{1}{M}\sum_{k=1}^{M}\frac{1}{n}\sum_{j=1}^{n} Match(C_j, L_j) \quad (5.6)$$

where $n$ is the number of instances. $C_j$ is classifier output label for feature $i$, $L_j$ is reference class label for feature $i$. When $C_j = L_j$, $Match$ is outputs 1 and, $C_j \neq L_j$, $Match$ is outputs 0.

- Average reduction amount of features

$$Reduction\ Rate = \frac{1}{M}\sum_{k=1}^{M} C - R_k \quad (5.7)$$

- Average computational time

$$AveCT = \frac{1}{M}\sum_{k=1}^{M} T_k \quad (5.8)$$

$T_k$ is the computational time for the $k_{th}$ run.

#### 5.1.4. Data description

Table 3 shows a list of machine learning datasets used in the computer experiments. These datasets are from the UCI Machine Learning Repository and the ASU Feature Selection Repository. They contain data collected from various fields such as medical, biology, and face images. In this paper, datasets No. 1 to No. 16 are categorized as small datasets and No. 17 to No. 26 are categorized as large datasets.

Table 3 : Datasets

| No. | Dataset | No.Features | No.Instances | No.Classes | Size |
|---|---|---|---|---|---|
| 1 | Breast-cancer | 9 | 699 | 2 | Small |
| 2 | Congress | 16 | 435 | 2 | |
| 3 | Ionosphere | 34 | 351 | 2 | |
| 4 | Sonar | 60 | 208 | 2 | |
| 5 | Statlog (Heart) | 13 | 270 | 2 | |
| 6 | Wine | 13 | 178 | 3 | |
| 7 | Zoo | 16 | 101 | 7 | |
| 8 | Waveform | 40 | 5000 | 3 | |
| 9 | SPECT Heart | 22 | 267 | 2 | |
| 10 | Lymphography | 18 | 148 | 4 | |
| 11 | Tic-Tac-Toe | 9 | 958 | 2 | |
| 12 | Exactly | 13 | 1000 | 2 | |
| 13 | Exactly2 | 13 | 1000 | 2 | |
| 14 | M-of-n | 13 | 1000 | 2 | |
| 15 | KrVsKp | 36 | 3196 | 2 | |
| 16 | Vote | 16 | 300 | 2 | |
| 17 | musk-1 | 166 | 476 | 2 | Large |
| 18 | Penglung | 325 | 73 | 7 | |
| 19 | SCADI | 206 | 70 | 7 | |
| 20 | landcover | 148 | 168 | 9 | |
| 21 | Yale | 1024 | 165 | 15 | |
| 22 | Colon | 2000 | 62 | 2 | |
| 23 | Prostate_GE | 5966 | 102 | 2 | |
| 24 | leukemia | 7070 | 72 | 2 | |
| 25 | orlraws10P | 10304 | 100 | 10 | |
| 26 | CLL_SUB_111 | 11340 | 111 | 3 | |

5.1.5. Compared Algorithm

The comparison algorithms are the following seven types of algorithms, including the population-based algorithm widely used in discrete problems and the swarm intelligence algorithm extended to discrete types.

- Genetic Algorithm (GA)
- Binary Particle Swarm Optimization (BPSO)
- Binary Ant Colony Optimization (BACO)
- Binary Grey Wolf Optimization (BGWO)
- Binary Whale Optimization Algorithm (BWOA)
- Binary Dragonfly Algorithm (BDA)
- Binary Harris Hawks Optimization (BHHO)

5.1.6. Parameter setting

Table 4 shows the parameter settings for each algorithm, and Table 5 shows the parameter settings for the proposed and comparison algorithm.

Table 4 : Parameter setting for each algorithm

| Run times $M$ | 30 |
|---|---|
| MaxIteration | 100 |
| Population number $N$ | 40 |

Table 5 : Parameter for the comparison algorithm.

| Algorithm | Parameter | Value |
|---|---|---|
| GA | Crossover rate | 0.8 |
| | Mutation rate | 0.3 |
| PSO | Inertia weight | 1 |
| | Cognitive factor | 2 |
| | Social factor | 2 |
| ACO | Coefficient control $\tau$ | 1 |
| | Coefficient control $\eta$ | 0.1 |
| | Initial $\tau$ | 1 |
| | Initial $\eta$ | 1 |
| | Pheromone evaporation | 0.2 |
| FSRO | Max Distance | 80 |
| | Decision Distance | 6 |
| | $w_1$ | 0.75 |
| | $w_2$ | 1 |
| | $d_1$ | 40 |
| | $d_2$ | 20 |

5.1.7. Experimental environment

This experiment is executed on a computer with Apple M1 CPU, 16.0GB RAM, macOS Ventura 13.4.1 operating system. All algorithms are coded in MATLAB 2023b.

5.2. Experimental result

Table 6 to Table 13 show the result of the mean, best, worst, and standard deviation of the fitness value. The Wilcoxon signed rank test at a 5% significant level is adopted to evaluate the statistical significance of the differences between FSRO versus comparison algorithms. Wilcoxon signed rank test is a test method to check if there is a significant difference between two results. In the table, "-" indicates no significant difference, meaning that the performance of the two algorithms is competitive. "+" indicates a significant difference, meaning that there is a difference in performance between two algorithms.

Table 14 shows the average accuracy. Table 15 shows the results of the Wilcoxon signed rank test for average accuracy. Table 16 shows the average reduction amount of features. Table 17 shows the average computational time. The best values among these algorithms are shown in bold.

Table 6: Mean of fitness

| No. | Dataset | **FSRO** | GA | BPSO | BACO | BGWO | BWOA | BDA | BHHO |
|---|---|---|---|---|---|---|---|---|---|
| 1 | Breastcancer | 0.0519 | 0.0529 | 0.0520 | **0.0472** | 0.0494 | 0.0520 | 0.0493 | 0.0481 |
| 2 | Congressional | 0.0496 | 0.0415 | 0.0454 | **0.0403** | 0.0556 | 0.0449 | 0.0448 | 0.0439 |
| 3 | Ionosphere | 0.0963 | 0.0776 | 0.0777 | **0.0521** | 0.1122 | 0.0712 | 0.0560 | 0.0890 |
| 4 | Sonar | 0.1045 | 0.0812 | 0.0728 | 0.0845 | 0.1273 | 0.0992 | **0.0717** | 0.0957 |
| 5 | Statlog (Heart) | 0.1368 | 0.1382 | 0.1451 | 0.1488 | 0.1556 | 0.1474 | **0.1342** | 0.1378 |
| 6 | Wine | 0.0401 | 0.0363 | 0.0393 | **0.0362** | 0.0603 | 0.0573 | 0.0451 | 0.0423 |
| 7 | Zoo | 0.0559 | 0.0557 | 0.0457 | 0.0460 | 0.0623 | 0.0647 | **0.0394** | 0.0455 |
| 8 | Waveform | 0.1950 | 0.1861 | 0.1842 | 0.1997 | 0.1962 | 0.1996 | **0.1740** | 0.1975 |
| 9 | SPECT Heart | 0.1944 | 0.1877 | 0.1928 | 0.1877 | 0.1915 | 0.1836 | **0.1782** | 0.1904 |
| 10 | Lymphography | 0.0923 | **0.0865** | 0.1096 | 0.1062 | 0.1186 | 0.1382 | 0.0879 | 0.1108 |
| 11 | Tic-Tac-Toe | 0.2361 | **0.2284** | 0.2452 | 0.2344 | 0.2409 | 0.2416 | 0.2349 | 0.2318 |
| 12 | Exactly | 0.0510 | 0.0470 | 0.1077 | 0.0604 | 0.0950 | 0.1650 | **0.0462** | 0.0466 |
| 13 | Exactly2 | 0.2241 | 0.2261 | **0.2195** | 0.2268 | 0.2282 | 0.2327 | 0.2234 | 0.2294 |
| 14 | M-of-n | 0.0490 | **0.0462** | 0.0478 | 0.0562 | 0.0590 | 0.0771 | **0.0462** | 0.0477 |
| 15 | KrVsKp | 0.0697 | 0.0597 | 0.0662 | 0.0731 | 0.0750 | 0.0733 | **0.0498** | 0.0702 |
| 16 | Vote | 0.0417 | 0.0394 | 0.0449 | 0.0440 | 0.0580 | 0.0427 | **0.0392** | 0.0462 |
| 17 | musk-1 | 0.1059 | 0.0931 | 0.0806 | 0.0948 | 0.1088 | 0.0916 | **0.0755** | 0.1030 |
| 18 | Penglung | 0.0996 | 0.0770 | 0.1190 | **0.0486** | 0.1451 | 0.0512 | 0.0636 | 0.0882 |
| 19 | SCADI | 0.1031 | 0.1177 | 0.1224 | **0.0599** | 0.1340 | 0.0897 | 0.0885 | 0.1322 |
| 20 | landcover | 0.3186 | 0.3022 | 0.3637 | **0.1927** | 0.3744 | 0.1948 | 0.2799 | 0.3205 |
| 21 | Yale | 0.3106 | 0.3293 | 0.3323 | **0.2843** | 0.3373 | 0.2956 | 0.2927 | 0.3491 |
| 22 | Colon | 0.1596 | 0.1240 | 0.1776 | **0.0279** | 0.1721 | 0.0304 | 0.1332 | 0.1503 |
| 23 | Prostate_GE | 0.1139 | 0.1271 | 0.1520 | 0.0539 | 0.1505 | **0.0337** | 0.1299 | 0.1160 |
| 24 | leukemia | 0.0823 | 0.0842 | 0.1057 | 0.0209 | 0.1035 | **0.0046** | 0.0871 | 0.0729 |
| 25 | orlraws10P | 0.1351 | 0.1275 | 0.1603 | **0.0157** | 0.1199 | 0.0168 | 0.1264 | 0.0944 |
| 26 | CLL_SUB_111 | 0.2718 | 0.2479 | 0.2655 | 0.1599 | 0.3273 | **0.1234** | 0.2523 | 0.2651 |

Table 8: Best of fitness

| No. | Dataset | **FSRO** | GA | BPSO | BACO | BGWO | BWOA | BDA | BHHO |
|---|---|---|---|---|---|---|---|---|---|
| 1 | Breastcancer | 0.0287 | 0.0330 | 0.0352 | **0.0222** | 0.0287 | 0.0352 | 0.0287 | 0.0352 |
| 2 | Congressional | 0.0188 | **0.0166** | **0.0166** | **0.0166** | 0.0250 | **0.0166** | **0.0166** | **0.0166** |
| 3 | Ionosphere | 0.0393 | 0.0393 | 0.0334 | 0.0217 | 0.0669 | 0.0316 | **0.0118** | 0.0404 |
| 4 | Sonar | 0.0450 | 0.0467 | 0.0300 | **0.0217** | 0.0567 | 0.0333 | **0.0217** | 0.0450 |
| 5 | Statlog (Heart) | 0.0974 | 0.0718 | 0.0808 | 0.0808 | **0.0564** | 0.0885 | 0.0962 | 0.0718 |
| 6 | Wine | **0.0154** | **0.0154** | **0.0154** | **0.0154** | 0.0231 | 0.0231 | **0.0154** | **0.0154** |
| 7 | Zoo | 0.0188 | **0.0125** | **0.0125** | **0.0125** | 0.0188 | **0.0125** | **0.0125** | 0.0188 |
| 8 | Waveform | 0.1799 | 0.1776 | 0.1568 | 0.1770 | 0.1745 | 0.1772 | **0.1520** | 0.1828 |
| 9 | SPECT Heart | 0.1043 | **0.0555** | 0.1064 | 0.1064 | 0.1110 | 0.1064 | 0.0873 | 0.1031 |
| 10 | Lymphography | 0.0333 | 0.0333 | 0.0222 | 0.0500 | 0.0500 | 0.0644 | **0.0167** | 0.0278 |
| 11 | Tic-Tac-Toe | 0.1969 | 0.2016 | 0.1990 | 0.2063 | **0.1942** | **0.1942** | 0.2016 | 0.2030 |
| 12 | Exactly | **0.0462** | **0.0462** | **0.0462** | **0.0462** | **0.0462** | **0.0462** | **0.0462** | **0.0462** |
| 13 | Exactly2 | 0.1832 | 0.1967 | **0.1562** | **0.1562** | 0.1864 | 0.1787 | 0.1742 | 0.1832 |
| 14 | M-of-n | **0.0462** | **0.0462** | **0.0462** | **0.0462** | **0.0462** | **0.0462** | **0.0462** | **0.0462** |
| 15 | KrVsKp | 0.0586 | 0.0460 | 0.0447 | 0.0614 | 0.0572 | 0.0531 | **0.0419** | 0.0544 |
| 16 | Vote | 0.0125 | **0.0063** | 0.0125 | 0.0188 | 0.0250 | 0.0125 | 0.0188 | 0.0125 |
| 17 | musk-1 | 0.0541 | 0.0577 | **0.0480** | 0.0491 | 0.0798 | 0.0585 | 0.0498 | 0.0491 |
| 18 | Penglung | 0.0363 | 0.0348 | 0.0298 | 0.0049 | 0.0462 | **0.0034** | 0.0231 | 0.0234 |
| 19 | SCADI | 0.0356 | 0.0346 | 0.0263 | 0.0020 | 0.0424 | **0.0010** | 0.0190 | 0.0215 |
| 20 | landcover | 0.2031 | 0.2119 | 0.1963 | 0.1125 | 0.2003 | **0.0579** | 0.1383 | 0.1615 |
| 21 | Yale | 0.1906 | 0.2089 | 0.2313 | 0.1732 | 0.2225 | **0.1688** | 0.2063 | 0.2423 |
| 22 | Colon | 0.0449 | 0.0443 | 0.0422 | 0.0004 | 0.0466 | **0.0001** | 0.0393 | 0.0244 |
| 23 | Prostate_GE | 0.0471 | 0.0470 | 0.0449 | 0.0013 | 0.0484 | **0.0001** | 0.0420 | 0.0245 |
| 24 | leukemia | 0.0470 | 0.0471 | 0.0450 | 0.0004 | 0.0481 | **0.0000** | 0.0359 | 0.0248 |
| 25 | orlraws10P | 0.0476 | 0.0475 | 0.0906 | 0.0004 | 0.0487 | **0.0000** | 0.0424 | 0.0241 |
| 26 | CLL_SUB_111 | 0.1309 | 0.1513 | 0.1284 | 0.0416 | 0.1859 | **0.0012** | 0.1287 | 0.1199 |

Table 7: Wilcoxon singed rank test of mean of fitness

| FSRO vs | Datasets | p-value | Decision | FSRO vs | Datasets | p-value | Decision |
|---|---|---|---|---|---|---|---|
| GA | Small | 0.6156 | - | BWOA | Small | 0.4016 | - |
|  | Large | 0.8534 | - |  | Large | 0.0355 | + |
|  | Overall | 0.6565 | - |  | Overall | 0.3126 | - |
| BPSO | Small | 0.9260 | - | BDA | Small | 0.3364 | - |
|  | Large | 0.4813 | - |  | Large | 0.4813 | - |
|  | Overall | 0.8065 | - |  | Overall | 0.3656 | - |
| BACO | Small | 0.8381 | - | BHHO | Small | 0.7240 | - |
|  | Large | 0.0433 | + |  | Large | 0.8534 | - |
|  | Overall | 0.1018 | - |  | Overall | 0.7785 | - |
| BGWO | Small | 0.3045 | - |  |  |  |  |
|  | Large | 0.2475 | - |  |  |  |  |
|  | Overall | 0.2788 | - |  |  |  |  |

Table 9: Wilcoxon singed rank test of best of fitness

| FSRO vs | Datasets | p-value | Decision | FSRO vs | Datasets | p-value | Decision |
|---|---|---|---|---|---|---|---|
| GA | Small | 0.8306 | - | BWOA | Small | 0.9926 | - |
|  | Large | 1.0000 | - |  | Large | 0.0334 | + |
|  | Overall | 0.8881 | - |  | Overall | 0.0749 | - |
| BPSO | Small | 0.6753 | - | BDA | Small | 0.4278 | - |
|  | Large | 0.8113 | - |  | Large | 0.3930 | - |
|  | Overall | 0.5580 | - |  | Overall | 0.2766 | - |
| BACO | Small | 0.8890 | - | BHHO | Small | 0.9778 | - |
|  | Large | 0.0430 | + |  | Large | 0.1655 | - |
|  | Overall | 0.1045 | - |  | Overall | 0.3952 | - |
| BGWO | Small | 0.5448 | - |  |  |  |  |
|  | Large | 0.4813 | - |  |  |  |  |
|  | Overall | 0.3417 | - |  |  |  |  |

As a result of the mean of fitness, the proposed algorithm did not show the best value on all datasets. However, Table 7 shows that the FSRO is a competitive algorithm on small datasets, since it does not have significant differences from all comparison algorithms. As the number of features increases, the difference in mean of fitness between the FSRO and the comparison algorithms increases, and a significant difference between the FSRO and BACO and BWOA is identified for large datasets. Therefore, it was found that the proposed algorithm has poor search performance for high dimensional problems.

As a result of the best of fitness, the proposed algorithm shows the best values for the three datasets Wine, Exactly, and M-of-n. Table 9 shows that the proposed algorithm is competitive on small datasets since there is no significant difference between the FSRO and all the comparison algorithms. However, it was found that with the increase in features in large-scale datasets, there is a difference of approximately 10 to 100 times. Significant differences were observed between FSRO and BACO and BWOA.

The results show that convergence speed was slow and performance of exploitation was low for high-dimensional problems.

Table 10: Worst of fitness

| No. | Dataset | FSRO | GA | BPSO | BACO | BGWO | BWOA | BDA | BHHO |
|---|---|---|---|---|---|---|---|---|---|
| 1 | Breastcancer | 0.0787 | **0.0673** | 0.0787 | 0.0787 | 0.0722 | 0.0768 | 0.0675 | 0.0675 |
| 2 | Congressional | 0.0974 | 0.0726 | 0.0912 | 0.0746 | 0.0830 | 0.0787 | 0.0830 | **0.0705** |
| 3 | Ionosphere | 0.2006 | 0.1036 | 0.1323 | **0.0830** | 0.1687 | 0.1146 | 0.1036 | 0.1866 |
| 4 | Sonar | 0.1903 | 0.1481 | 0.1464 | **0.1364** | 0.2017 | 0.1567 | 0.1584 | 0.1870 |
| 5 | Statlog (Heart) | **0.1897** | 0.2051 | 0.2474 | 0.1962 | 0.2628 | 0.1987 | 0.1974 | 0.2064 |
| 6 | Wine | 0.0745 | **0.0642** | 0.1079 | 0.0925 | 0.1233 | 0.0976 | 0.0822 | 0.1079 |
| 7 | Zoo | 0.1663 | 0.1987 | 0.0888 | 0.1525 | 0.1788 | 0.1600 | **0.0825** | 0.1013 |
| 8 | Waveform | 0.2062 | 0.1966 | 0.1984 | 0.2138 | 0.2196 | 0.2166 | **0.1910** | 0.2087 |
| 9 | SPECT Heart | 0.2932 | 0.2774 | 0.2774 | 0.2932 | 0.2696 | **0.2593** | 0.2729 | 0.3205 |
| 10 | Lymphography | 0.1830 | **0.1464** | 0.2195 | 0.1774 | 0.1996 | 0.1996 | 0.1519 | 0.1996 |
| 11 | Tic-Tac-Toe | 0.2696 | 0.2723 | 0.3006 | 0.2689 | 0.3006 | 0.2932 | 0.2723 | **0.2555** |
| 12 | Exactly | 0.0763 | 0.0642 | 0.2912 | 0.1470 | 0.2815 | 0.3092 | **0.0462** | 0.0583 |
| 13 | Exactly2 | 0.2738 | 0.2886 | 0.2687 | 0.2796 | 0.2912 | 0.2777 | 0.2867 | **0.2642** |
| 14 | M-of-n | 0.0615 | **0.0462** | 0.0583 | 0.0840 | 0.1367 | 0.1645 | **0.0462** | 0.0538 |
| 15 | KrVsKp | 0.0866 | 0.0728 | 0.0953 | 0.0855 | 0.0920 | 0.0993 | **0.0669** | 0.0810 |
| 16 | Vote | **0.0700** | 0.0763 | 0.1063 | 0.0813 | 0.0950 | 0.0763 | 0.0763 | 0.0975 |
| 17 | musk-1 | 0.1595 | 0.1482 | **0.1091** | 0.1578 | 0.1485 | 0.1381 | 0.1216 | 0.1498 |
| 18 | Penglung | 0.3051 | 0.2362 | 0.3577 | 0.1953 | 0.3095 | **0.1360** | 0.1609 | 0.1587 |
| 19 | SCADI | 0.2338 | 0.2372 | 0.2236 | 0.1997 | 0.3006 | **0.1992** | 0.2757 | 0.2815 |
| 20 | landcover | 0.4267 | 0.4233 | 0.4731 | **0.2870** | 0.4921 | 0.3307 | 0.3538 | 0.4301 |
| 21 | Yale | 0.4034 | 0.4851 | 0.4498 | 0.4190 | 0.4222 | 0.4084 | **0.3997** | 0.4539 |
| 22 | Colon | 0.3464 | 0.2726 | 0.4177 | **0.0799** | 0.3500 | 0.2252 | 0.3403 | 0.4095 |
| 23 | Prostate_GE | 0.2288 | 0.2274 | 0.3600 | **0.0955** | 0.3704 | 0.1353 | 0.2209 | 0.2048 |
| 24 | leukemia | 0.2419 | 0.1765 | 0.3025 | 0.1311 | 0.3697 | **0.0645** | 0.2355 | 0.1652 |
| 25 | orlraws10P | 0.3628 | 0.2734 | 0.3168 | **0.0460** | 0.2289 | 0.0902 | 0.2680 | 0.2076 |
| 26 | CLL_SUB_111 | 0.4184 | 0.3565 | 0.4972 | 0.2869 | 0.5548 | **0.2048** | 0.3745 | 0.3657 |

Table 11: Wilcoxon singed rank test of worst of fitness

| FSRO vs | Datasets | p-value | Decision | FSRO vs | Datasets | p-value | Decision |
|---|---|---|---|---|---|---|---|
| GA | Small | 0.4967 | - | BWOA | Small | 0.4505 | - |
| | Large | 0.5288 | - | | Large | 0.0089 | + |
| | Overall | 0.4426 | - | | Overall | 0.3147 | - |
| BPSO | Small | 0.4971 | - | BDA | Small | 0.4284 | - |
| | Large | 0.4813 | - | | Large | 0.3930 | - |
| | Overall | 0.4872 | - | | Overall | 0.3903 | - |
| BACO | Small | 0.9482 | - | BHHO | Small | 0.9556 | - |
| | Large | 0.0185 | + | | Large | 0.5787 | - |
| | Overall | 0.1404 | - | | Overall | 0.6305 | - |
| BGWO | Small | 0.2745 | - | | | | |
| | Large | 0.4813 | - | | | | |
| | Overall | 0.3419 | - | | | | |

The results of the worst of fitness show that the proposed algorithm performs best on two datasets, Statlog (Heart) and Vote. Since these are small datasets, it was found that the proposed algorithm performs better in the worst value with small dimension problems. Table 11 shows that significant differences were only found for the BACO and BWOA high-dimensional datasets. However, the algorithm has the worst value for some datasets, regardless of the size of the problem, indicating that this is the reason why the FSRO does not perform the best on the mean of fitness.

Table 12: Standard deviation of fitness

| No. | Dataset | FSRO | GA | BPSO | BACO | BGWO | BWOA | BDA | BHHO |
|---|---|---|---|---|---|---|---|---|---|
| 1 | Breastcancer | 0.0113 | 0.0109 | 0.0101 | 0.0138 | 0.0103 | 0.0109 | **0.0095** | 0.0096 |
| 2 | Congressional | 0.0181 | 0.0161 | 0.0187 | 0.0155 | 0.0154 | 0.0209 | 0.0158 | **0.0137** |
| 3 | Ionosphere | 0.0315 | 0.0161 | 0.0217 | **0.0153** | 0.0274 | 0.0247 | 0.0212 | 0.0325 |
| 4 | Sonar | 0.0361 | **0.0247** | 0.0305 | 0.0309 | 0.0377 | 0.0379 | 0.0322 | 0.0372 |
| 5 | Statlog (Heart) | **0.0236** | 0.0374 | 0.0350 | 0.0249 | 0.0423 | 0.0304 | 0.0246 | 0.0340 |
| 6 | Wine | 0.0163 | **0.0159** | 0.0201 | 0.0179 | 0.0251 | 0.0221 | 0.0193 | 0.0263 |
| 7 | Zoo | 0.0364 | 0.0473 | 0.0249 | 0.0327 | 0.0342 | 0.0404 | **0.0202** | 0.0252 |
| 8 | Waveform | 0.0070 | **0.0059** | 0.0102 | 0.0079 | 0.0097 | 0.0085 | 0.0091 | 0.0066 |
| 9 | SPECT Heart | 0.0478 | 0.0530 | 0.0481 | 0.0473 | **0.0352** | 0.0441 | 0.0434 | 0.0496 |
| 10 | Lymphography | 0.0397 | **0.0281** | 0.0468 | 0.0320 | 0.0336 | 0.0361 | 0.0322 | 0.0421 |
| 11 | Tic-Tac-Toe | 0.0188 | 0.0151 | 0.0215 | 0.0157 | 0.0241 | 0.0252 | 0.0176 | **0.0149** |
| 12 | Exactly | 0.0089 | 0.0035 | 0.0981 | 0.0234 | 0.0692 | 0.1040 | **0.0000** | 0.0022 |
| 13 | Exactly2 | 0.0259 | 0.0206 | 0.0261 | 0.0289 | 0.0297 | 0.0233 | 0.0266 | **0.0198** |
| 14 | M-of-n | 0.0043 | **0.0000** | 0.0035 | 0.0098 | 0.0180 | 0.0371 | **0.0000** | 0.0031 |
| 15 | KrVsKp | 0.0062 | 0.0061 | 0.0108 | 0.0061 | 0.0077 | 0.0105 | **0.0052** | 0.0065 |
| 16 | Vote | 0.0156 | 0.0159 | 0.0201 | 0.0173 | 0.0196 | 0.0165 | **0.0145** | 0.0203 |
| 17 | musk-1 | 0.0231 | 0.0218 | 0.0159 | 0.0230 | 0.0179 | 0.0216 | **0.0153** | 0.0219 |
| 18 | Penglung | 0.0712 | 0.0498 | 0.0723 | 0.0482 | 0.0716 | 0.0472 | **0.0433** | 0.0458 |
| 19 | SCADI | 0.0584 | 0.0559 | 0.0666 | **0.0536** | 0.0598 | 0.0614 | 0.0600 | 0.0751 |
| 20 | landcover | 0.0575 | **0.0481** | 0.0731 | 0.0491 | 0.0675 | 0.0648 | 0.0557 | 0.0591 |
| 21 | Yale | 0.0543 | 0.0666 | 0.0611 | 0.0690 | 0.0613 | 0.0613 | **0.0497** | 0.0634 |
| 22 | Colon | 0.0758 | 0.0721 | 0.0971 | **0.0354** | 0.0876 | 0.0505 | 0.0825 | 0.1007 |
| 23 | Prostate_GE | 0.0565 | 0.0497 | 0.0848 | **0.0283** | 0.0757 | 0.0368 | 0.0544 | 0.0543 |
| 24 | leukemia | 0.0501 | 0.0467 | 0.0714 | 0.0423 | 0.0756 | **0.0163** | 0.0511 | 0.0522 |
| 25 | orlraws10P | 0.0688 | 0.0665 | 0.0677 | **0.0185** | 0.0509 | 0.0277 | 0.0602 | 0.0529 |
| 26 | CLL_SUB_111 | 0.0734 | 0.0582 | 0.0867 | 0.0559 | 0.0817 | **0.0473** | 0.0612 | 0.0626 |

Table 13: Wilcoxon singed rank test of std of fitness

| FSRO vs | Datasets | p-value | Decision | FSRO vs | Datasets | p-value | Decision |
|---|---|---|---|---|---|---|---|
| GA | Small | 0.5087 | - | BWOA | Small | 0.1997 | - |
| | Large | 0.1903 | - | | Large | 0.0524 | - |
| | Overall | 0.4928 | - | | Overall | 0.9747 | - |
| BPSO | Small | 0.4848 | - | BDA | Small | 0.6090 | - |
| | Large | 0.1051 | - | | Large | 0.3930 | - |
| | Overall | 0.3953 | - | | Overall | 0.5520 | - |
| BACO | Small | 0.8672 | - | BHHO | Small | 0.9556 | - |
| | Large | 0.0115 | + | | Large | 0.6987 | - |
| | Overall | 0.3126 | - | | Overall | 0.8811 | - |
| BGWO | Small | 0.3226 | - | | | | |
| | Large | 0.2475 | - | | | | |
| | Overall | 0.3681 | - | | | | |

From the results of the standard deviation of fitness, the proposed method showed the best values in the Statlog (Heart) dataset. The standard deviation in other datasets is at most 0.0758 for Colon, which is lower compared to the maximum values of the comparison algorithms. Table 13 shows that only BACO is significantly different from the proposed algorithm, indicating its superior performance in terms of standard deviation. However, the mean of fitness did not show the best value, indicating that the quality of the fitness value is stable at a low level.

Table 14: Average accuracy

| No. | Dataset | FSRO | GA | BPSO | BACO | BGWO | BWOA | BDA | BHHO |
|---|---|---|---|---|---|---|---|---|---|
| 1 | Breastcancer | 0.9736 | 0.9746 | 0.9739 | 0.9767 | **0.9772** | 0.9727 | 0.9765 | 0.9758 |
| 2 | Congressional | **0.9720** | 0.9693 | 0.9701 | 0.9670 | 0.9678 | 0.9605 | 0.9655 | 0.9709 |
| 3 | Ionosphere | 0.9319 | 0.9400 | 0.9424 | 0.9533 | 0.9243 | 0.9329 | **0.9614** | 0.9262 |
| 4 | Sonar | 0.9334 | **0.9561** | **0.9561** | 0.9301 | 0.9187 | 0.9130 | 0.9496 | 0.9374 |
| 5 | Statlog (Heart) | 0.8871 | 0.8815 | 0.8784 | 0.8685 | 0.8741 | 0.8673 | **0.8877** | 0.8833 |
| 6 | Wine | **0.9925** | 0.9905 | 0.9905 | 0.9886 | 0.9771 | 0.9657 | 0.9838 | 0.9838 |
| 7 | Zoo | 0.9840 | 0.9733 | 0.9817 | 0.9867 | 0.9750 | 0.9617 | **0.9900** | 0.9867 |
| 8 | Waveform | 0.8359 | 0.8401 | 0.8441 | 0.8236 | **0.8503** | 0.8235 | 0.8502 | 0.8361 |
| 9 | SPECT Heart | 0.8141 | 0.8088 | 0.7994 | 0.8000 | 0.8101 | 0.8019 | **0.8220** | 0.8044 |
| 10 | Lymphography | **0.9485** | 0.9425 | 0.9161 | 0.9207 | 0.9184 | 0.8839 | 0.9356 | 0.9230 |
| 11 | Tic-Tac-Toe | 0.8118 | 0.8260 | 0.7976 | 0.8169 | 0.8105 | **0.8271** | 0.8213 | 0.8248 |
| 12 | Exactly | 0.9975 | 0.9993 | 0.9257 | 0.9902 | 0.9537 | 0.8565 | **1.0000** | 0.9998 |
| 13 | Exactly2 | 0.7641 | 0.7625 | **0.7647** | 0.7585 | 0.7598 | 0.7500 | 0.7643 | 0.7545 |
| 14 | M-of-n | **1.0000** | **1.0000** | 0.9998 | 0.9965 | 0.9937 | 0.9773 | **1.0000** | **1.0000** |
| 15 | KrVsKp | 0.9754 | 0.9743 | 0.9692 | 0.9607 | **0.9785** | 0.9587 | 0.9764 | 0.9728 |
| 16 | Vote | **0.9833** | 0.9778 | 0.9728 | 0.9717 | 0.9756 | 0.9667 | 0.9817 | 0.9711 |
| 17 | musk-1 | 0.9351 | 0.9488 | 0.9561 | 0.9214 | 0.9516 | 0.9270 | **0.9632** | 0.9312 |
| 18 | Penglung | 0.9396 | 0.9595 | 0.9048 | 0.9548 | 0.8952 | 0.9500 | **0.9619** | 0.9357 |
| 19 | SCADI | 0.9340 | 0.9119 | 0.8952 | **0.9405** | 0.9000 | 0.9048 | 0.9286 | 0.8857 |
| 20 | landcover | 0.6980 | 0.7162 | 0.6333 | **0.7970** | 0.6515 | 0.7919 | 0.7212 | 0.6949 |
| 21 | Yale | 0.7095 | 0.6869 | 0.6778 | 0.6985 | 0.6949 | 0.6889 | **0.7222** | 0.6586 |
| 22 | Colon | 0.8755 | 0.9139 | 0.8500 | **0.9722** | 0.8667 | 0.9667 | 0.8972 | 0.8867 |
| 23 | Prostate_GE | 0.9272 | 0.9117 | 0.8817 | 0.9433 | 0.8900 | **0.9633** | 0.9017 | 0.9033 |
| 24 | leukemia | 0.9626 | 0.9595 | 0.9333 | 0.9786 | 0.9405 | **0.9952** | 0.9500 | 0.9524 |
| 25 | orlraws10P | 0.9040 | 0.9117 | 0.8733 | **0.9833** | 0.9217 | 0.9817 | 0.9067 | 0.9250 |
| 26 | CLL_SUB_111 | 0.7534 | 0.7795 | 0.7576 | 0.8303 | 0.7061 | **0.8652** | 0.7712 | 0.7530 |

Table 15: Wilcoxon singed rank test of average accuracy

| FSRO vs | Datasets | p-value | Decision | FSRO vs | Datasets | p-value | Decision |
|---|---|---|---|---|---|---|---|
| GA | Small | 0.9344 | - | BWOA | Small | 0.1596 | - |
|  | Large | 0.8973 | - |  | Large | 0.2799 | - |
|  | Overall | 0.8953 | - |  | Overall | 0.7199 | - |
| BPSO | Small | 0.5147 | - | BDA | Small | 0.7803 | - |
|  | Large | 0.3527 | - |  | Large | 0.7959 | - |
|  | Overall | 0.4590 | - |  | Overall | 0.7371 | - |
| BACO | Small | 0.6963 | - | BHHO | Small | 0.7885 | - |
|  | Large | 0.1655 | - |  | Large | 0.5787 | - |
|  | Overall | 0.6599 | - |  | Overall | 0.6732 | - |
| BGWO | Small | 0.4909 | - |  |  |  |  |
|  | Large | 0.4359 | - |  |  |  |  |
|  | Overall | 0.4508 | - |  |  |  |  |

The results of accuracy showed that the proposed algorithm obtained the best values on the five datasets of Congressional Voting Records, Wine, Lymphography, M-of-n, and Vote, and showed 100% classification accuracy in M-of-n. The average accuracy for all small datasets was 92.53% and for all large datasets was 86.39%. The results of the large datasets did not show the best accuracy, with landcover, Colon, and CLL_SUB_111 showing a difference of about 10% from the comparison algorithms.

However, as shown in Table 15, the proposed algorithm did not show any significant difference compared to the comparison algorithms in all test results, indicating that it has a competitive performance in terms of accuracy.

Table 16: Average reduction amount of features

| No. | Dataset | FSRO | GA | BPSO | BACO | BGWO | BWOA | BDA | BHHO |
|---|---|---|---|---|---|---|---|---|---|
| 1 | Breastcancer | 6.5 | 6.3 | 6.4 | **6.6** | 6.4 | 6.5 | 6.5 | **6.6** |
| 2 | Congressional | 12.1 | 13.8 | 13.0 | 14.3 | 11.7 | **14.5** | 13.8 | 13.2 |
| 3 | Ionosphere | 22.1 | 26.0 | 25.2 | **30.6** | 19.0 | 30.3 | 26.8 | 26.3 |
| 4 | Sonar | 33.2 | 35.0 | 40.0 | 47.0 | 27.5 | **47.4** | 44.2 | 36.4 |
| 5 | Statlog (Heart) | 8.4 | 8.9 | 8.4 | 9.0 | 7.5 | **9.4** | 8.7 | 8.7 |
| 6 | Wine | 8.7 | 9.4 | 9.0 | **9.6** | 7.8 | 9.6 | 9.0 | 9.4 |
| 7 | Zoo | 9.4 | 10.9 | **11.3** | 10.6 | 9.6 | 11.2 | 11.1 | 10.6 |
| 8 | Waveform | 21.1 | 23.1 | 22.5 | 23.6 | 15.4 | 23.7 | **24.3** | 20.0 |
| 9 | SPECT Heart | 16.0 | 18.6 | 19.3 | 20.3 | 17.5 | **20.8** | 18.0 | 18.8 |
| 10 | Lymphography | 9.7 | 11.7 | 11.9 | 11.7 | 9.9 | 11.9 | **12.6** | 10.5 |
| 11 | Tic-Tac-Toe | 3.0 | 2.5 | **3.3** | 2.7 | 2.7 | 1.3 | 2.3 | 2.3 |
| 12 | Exactly | 6.7 | 7.0 | 7.7 | 6.3 | 6.1 | **8.3** | 7.0 | 7.0 |
| 13 | Exactly2 | 11.5 | 11.4 | **12.0** | 11.8 | 11.4 | **12.0** | 11.5 | 11.9 |
| 14 | M-of-n | 5.8 | **7.0** | 6.8 | 6.1 | 6.1 | 5.6 | **7.0** | 6.8 |
| 15 | KrVsKp | 18.9 | 22.8 | 22.2 | 22.4 | 15.8 | 23.0 | **25.7** | 19.5 |
| 16 | Vote | 11.7 | 12.9 | 12.7 | 13.0 | 10.2 | **14.0** | 12.4 | 12.8 |
| 17 | musk-1 | 87.1 | 88.1 | 97.7 | **126.1** | 57.8 | 122.9 | 95.7 | 97.7 |
| 18 | Penglung | 178.1 | 193.1 | 216.9 | 299.5 | 159.8 | **304.7** | 229.8 | 226.2 |
| 19 | SCADI | 116.5 | 127.3 | 148.4 | 193.1 | 115.8 | **197.9** | 156.4 | 145.8 |
| 20 | landcover | 79.2 | 79.3 | 98.5 | 133.4 | 58.6 | **136.9** | 105.4 | 80.5 |
| 21 | Yale | 520.3 | 537.7 | 591.0 | **891.5** | 381.3 | 863.9 | 586.8 | 595.4 |
| 22 | Colon | 1049.2 | 1069.0 | 1147.5 | 1942.7 | 957.2 | **1992.3** | 1185.6 | 1393.3 |
| 23 | Prostate_GE | 3078.5 | 3127.2 | 3254.0 | 5791.0 | 2891.7 | **5921.7** | 3494.3 | 4234.9 |
| 24 | leukemia | 3633.7 | 3691.3 | 3840.0 | 6958.6 | 3537.4 | **7050.0** | 4093.4 | 4949.3 |
| 25 | orlraws10P | 5283.5 | 5356.2 | 5537.6 | 10226.8 | 5209.7 | **10272.0** | 5935.9 | 7531.6 |
| 26 | CLL_SUB_111 | 5684.9 | 5731.6 | 5972.6 | 10532.1 | 4224.5 | **11108.0** | 6077.1 | 6479.7 |

Table 17: Average computational time

| No. | Dataset | FSRO | GA | BPSO | BACO | BGWO | BWOA | BDA | BHHO |
|---|---|---|---|---|---|---|---|---|---|
| 1 | Breastcancer | 12.521 | 19.976 | 12.431 | **8.381** | 12.657 | 11.090 | 12.388 | 22.535 |
| 2 | Congressional | 11.744 | 18.597 | 11.815 | 11.485 | 11.420 | **9.225** | 11.646 | 20.084 |
| 3 | Ionosphere | **10.498** | 16.946 | 11.172 | 11.021 | 10.723 | 10.559 | 10.989 | 18.773 |
| 4 | Sonar | 12.560 | 19.902 | 12.800 | 14.922 | 13.500 | 13.873 | **12.463** | 21.977 |
| 5 | Statlog (Heart) | 11.568 | 17.638 | 11.006 | 12.159 | 11.854 | **10.565** | 10.961 | 20.741 |
| 6 | Wine | 10.390 | 16.615 | 10.688 | 11.792 | 10.794 | **10.262** | 10.747 | 19.980 |
| 7 | Zoo | 11.064 | 16.463 | 10.565 | 11.300 | 11.051 | **9.354** | 10.648 | 18.690 |
| 8 | Waveform | 87.238 | 130.395 | 78.280 | 92.337 | 93.851 | **76.687** | 79.652 | 171.360 |
| 9 | SPECT Heart | 11.032 | 17.578 | 11.249 | 12.061 | 11.683 | **10.289** | 11.197 | 19.076 |
| 10 | Lymphography | **9.941** | 17.370 | 10.922 | 10.858 | 10.849 | 10.153 | 10.706 | 18.862 |
| 11 | Tic-Tac-Toe | 13.593 | 20.396 | 13.596 | 14.954 | 14.736 | 14.973 | **13.382** | 24.125 |
| 12 | Exactly | 14.655 | 22.441 | 14.090 | 13.809 | 14.234 | **12.005** | 13.810 | 24.827 |
| 13 | Exactly2 | 14.316 | 21.375 | 12.946 | 13.444 | 12.716 | **10.463** | 12.965 | 24.617 |
| 14 | M-of-n | 14.278 | 22.391 | 14.245 | 13.496 | 14.762 | 14.406 | **13.198** | 25.277 |
| 15 | KrVsKp | 39.914 | 62.557 | 37.535 | 41.777 | 43.490 | **35.395** | 38.181 | 81.565 |
| 16 | Vote | 11.732 | 17.831 | 11.110 | 12.326 | 12.246 | **10.292** | 11.054 | 20.225 |
| 17 | musk-1 | 18.778 | 20.699 | **12.590** | 14.430 | 14.691 | 12.976 | 14.005 | 24.113 |
| 18 | Penglung | 10.587 | 16.858 | 10.492 | 17.858 | 11.854 | **10.332** | 12.214 | 19.429 |
| 19 | SCADI | 10.257 | 16.465 | **9.767** | 14.402 | 11.032 | 9.912 | 11.596 | 18.750 |
| 20 | landcover | **9.372** | 16.839 | 10.596 | 13.340 | 11.056 | 9.867 | 11.431 | 18.387 |
| 21 | Yale | 13.701 | 21.929 | 14.270 | 106.617 | 18.117 | **10.835** | 20.972 | 26.881 |
| 22 | Colon | 12.785 | 20.584 | 13.363 | 350.153 | 16.891 | **9.010** | 26.327 | 23.661 |
| 23 | Prostate_GE | 25.327 | 40.878 | 24.285 | 2390.600 | 36.661 | **11.278** | 66.624 | 44.869 |
| 24 | leukemia | 21.194 | 36.291 | 22.123 | 3193.100 | 35.080 | **9.820** | 75.030 | 40.230 |
| 25 | orlraws10P | 41.311 | 54.710 | 33.945 | 6622.200 | 52.813 | **12.343** | 108.557 | 62.196 |
| 26 | CLL_SUB_111 | 37.709 | 60.170 | 38.859 | 7511.100 | 62.630 | **14.082** | 120.022 | 73.288 |

The results of feature reduction showed that the proposed method did not obtain the best values on all datasets. BWOA reduced 197.90 features from 206 features in SCADI, resulting in an accuracy worse than the other algorithms. The proposed algorithm avoids excessive feature reduction and achieves high accuracy.

The results of average computation time showed that the proposed algorithm performed best for Ionosphere, Lymphography, and landcover. A significant increase in computation time due to the increase in the number of features was observed with BACO, but not with FSRO. Therefore, the computation time of the proposed algorithm is not affected by the size of the features, and the results can be obtained within a feasible time.

## 5.3. Evaluation of performance

Based on the experimental results, an evaluation and discussion of the performance of the proposed algorithm are described.

### 5.3.1. Overall search performance evaluation

It was found that the proposed method is superior to the average accuracy and best of fitness on the datasets. From the results of the standard deviation of fitness, it was found that there was little variation in the fitness values for each run, indicating that the proposed method has a robust performance. The results of the mean of fitness did not show the best value, but neither did it show the worst value, except for SPECT Heart. Therefore, it was found that the proposed algorithm is not a suitable algorithm for a specific problem, while it is a versatile algorithm that shows intermediate performance for many problems. Since the proposed method shows stable accuracy without excessive feature reduction, it was found to be a well-balanced search algorithm that can simultaneously improve accuracy and reduce data. However, the proposed algorithm has poor search performance to flexibly adapt to high-dimensional problems because the accuracy decreases for data sets with more than 2,000 features. One reason for this is the slow convergence.

Table 18 shows the results of increasing the number of iterations from 100 to 500 for landcover, which the proposed algorithm performs significantly worse in terms of accuracy.

Table 18: Average accuracy

| Iteration | meanFitness | Accuracy | Computational Time |
|---|---|---|---|
| 100 | 0.3186 | 0.6980 | 9.3722 |
| 200 | 0.3081 | 0.7112 | 20.6978 |
| 300 | 0.3006 | 0.7198 | 31.4853 |
| 400 | 0.2937 | 0.7260 | 41.6035 |
| 500 | 0.2875 | 0.7331 | 52.0108 |

The results show that the average accuracy and mean of fitness improve as the number of iterations is increased. Therefore, the convergence of the proposed algorithm is slow, and the search for updating the solutions ends prematurely with short iterations. It is expected that improving the convergence speed of proposed algorithm improves the performance of the FSRO.

### 5.3.2. Effects of successful predation

There are three main methods for updating the solution in the FSRO: two-point crossover between snakes, uniform crossover between frogs, and inversion of elements in the successful predation phase. The inversion operation is important since it modifies a part of the solution based on the information among different groups. Figure 7 shows the convergence graph of fitness value for the Ionosphere, a small dataset frequently used to evaluate feature selection performance. The red dots in Figure 7 indicate the timing of successful predation.

Figure 7: FSRO convergence graph of Ionosphere

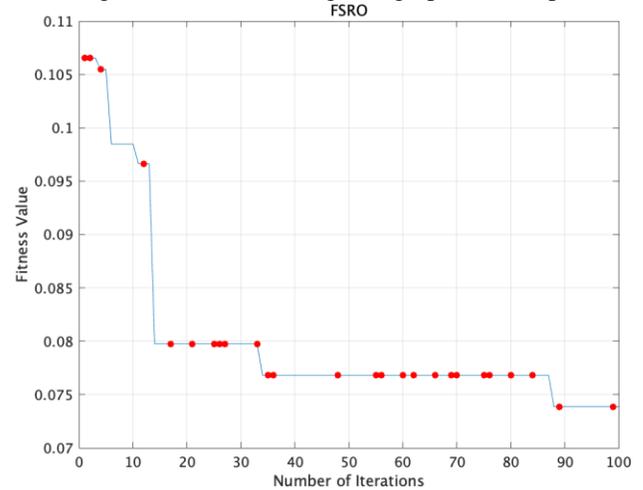

As shown in the figure, the better fitness value was updated when predation success at iteration=33. Since the fitness value was not updated in the previous 20 iterations, it indicates that the inversion of elements increases the diversity of the solution and is effective in avoiding the stagnation local optima.

Figure 8 shows the convergence graph of fitness value for the large dataset, CLL_SUB_111. Predation succeeds in the iterations before updating the solution, indicating that the inversion of elements is effective in improving the fitness value the same as in the small datasets. However, since predation does not succeed after the middle of the iteration, the inversion operation is not performed, and the improvement of the solution is stagnant.

Figure 8: FSRO convergence graph of CLL_SUB_111

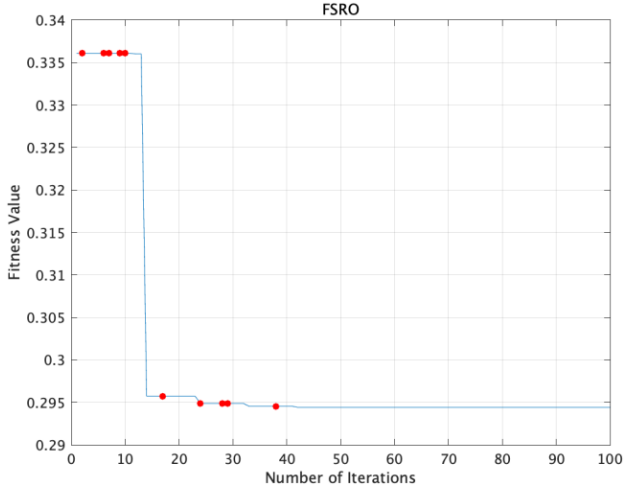

Figure 9 shows the variation of the frog-snake distance for the small and large datasets. One of the main reasons for predation not succeeding is that the variation of distance between frogs and snakes becomes small, leading to the execution of only some predation patterns. This distance is obtained based on the number of matched elements in the two solutions. Therefore, the frog-snake distance equation (4.3) needs to be modified considering the dimension of the problem.

Figure 9: Variations of frog-snake distance

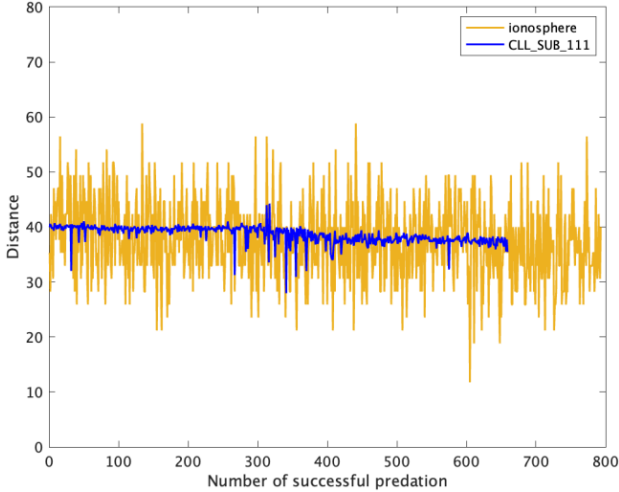

5.3.3.  Effects of introducing EGT

The proposed algorithm enables dynamic control of the search by introducing evolutionary game theory. Figure 9 shows the changes in the frog and snake population obtained in the same run as Figure 7. As previously mentioned, this run updates the better solution when predation successful at the iteration=33. Figure 10 shows a rapid increase in the frog population around the iteration=33.

Figure 10: Change of frog-snake population of Ionosphere

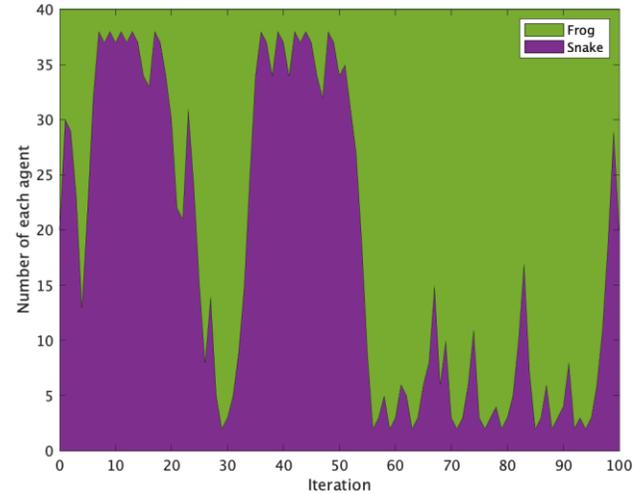

The snake group updates the solution by two-point crossover, which has the performance of exploration, indicating that the exploration is enhanced in FSRO after finding a better solution. Thus, after updating a better solution, the proposed algorithm prevents stagnation to a local optima and increases the possibility of updating a better solution by searching widely in the vicinity of the better solution. In addition, the number of frogs that update the solution by uniform crossover, which has exploitation, increases at the end of the iteration, indicating that the frogs are able to dynamically select an appropriate search method in a series of iterations. This indicates that the introduction of evolutionary game theory is effective in controlling dynamic search.

5.3.4.  Effects of mutation operation

When the population of one of the species accounts for the majority of the population, the proposed method uses mutation operations to maintain diversity in the search. If the number of population in a group becomes 2 or less, the best solution until current iteration is added to that group. Figure 11 shows the change in the number of populations of each groups in the same run as in Figure 9, but without the mutation operation.

As shown in the figure, after the number of snake populations becomes 2 or less, there is no change in the number of populations in subsequent iterations. This demonstrated that the diversity of search becomes low, and dynamic search becomes impossible. Therefore, the mutation operation is effective in maintaining the diversity of the search.

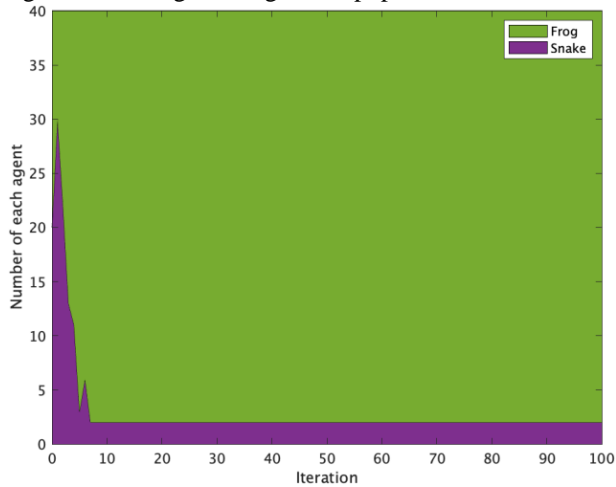

Figure 11: Change of frog-snake population without ESS

## 6. Conclusion and future work

This study proposed a novel swarm intelligence algorithm for applying binary optimization problems named Frog-Snake Prey-Predation Relationship Optimization (FSRO) inspired by the prey-predation relationship between frogs and snakes. The proposed algorithm models the search, approach, and capture behaviors of the snake, as well as the frog's characteristic behavior of staying still to attract and then escaping at each phases. The concepts of replicator dynamics and evolutionary stable strategy (ESS), which are important in evolutionary game theory (EGT), are introduced to control dynamic search and to improve search diversity.

Computational experiments were conducted on 26 machine learning datasets and the proposed algorithm was compared with seven binary versions of swarm intelligence algorithms. Experimental results showed that the proposed algorithm performed better than the comparison algorithm on some datasets in terms of the best and standard deviation values of fitness and average accuracy. It was found that the proposed algorithm is a well-balanced search algorithm that can simultaneously achieve the improvement of accuracy and reduction of data. The overall experimental results show that the proposed algorithm is not specific to any particular problem, but has an intermediate performance.

Solution updating during the predation phase was shown to be effective in increasing diversity and improving the quality of solutions. The introduction of evolutionary game theory enabled dynamic control of the timing of solution updating and the search method according to a series of iteration. The mutation operation was found to be an important operation in maintaining high search diversity and achieving dynamic search.

One of the improvements of the proposed method is the slow convergence. Since the search for updating the solutions ends prematurely with short iterations, it is necessary to limit the range of the uniform crossover and modify the equation for calculating the improvement of the solution. The reason for the inability to adapt flexibly to high-dimensional problems is that the distance variation between frogs and snakes decreases, and some predation patterns are not executed. Therefore, it is necessary to modify the distance calculation equation to increase the probability of successful predation.

Furthermore, in this study, computational experiments were conducted using datasets for performance testing. Therefore, future work is to evaluate the overall performance by applying to real-world problems. The proposed algorithm is designed to be applied to binary optimization problems. Therefore, it needs to be improved for application to integer optimization problems such as traveling salesman problems and scheduling problems, and its search concept and framework also need to be evaluated and discussed.

**Acknowledgements**

**Data availability**

**Funding**

**Declarations**
**Conflict of interest**

**Ethical approval**

**Informed consent**